\documentclass{article}
\usepackage{microtype}
\usepackage{graphicx}
\usepackage{subcaption}
\usepackage{booktabs}
\usepackage{hyperref}
\usepackage{listings}
\usepackage{xcolor}

\usepackage[preprint]{icml2026}
\usepackage{amsmath}
\usepackage{amssymb}
\usepackage{mathtools}
\usepackage{amsthm}

\usepackage[capitalize,noabbrev]{cleveref}

\theoremstyle{plain}

\theoremstyle{definition}

\theoremstyle{remark}

\usepackage[textsize=tiny]{todonotes}

\usepackage{times}
\usepackage{latexsym}
\usepackage{hyperref}
\usepackage{graphicx} 
\usepackage{url}
\usepackage{tcolorbox}
\usepackage[utf8]{inputenc} 
\usepackage[T1]{fontenc}    
\usepackage{hyperref}   
\usepackage{authblk}
\usepackage{wrapfig}
\usepackage{url}            
\usepackage{booktabs}       
\usepackage{amsfonts}       
\usepackage{nicefrac}       
\usepackage{microtype}      
\usepackage{xcolor}         
\usepackage{bm}
\usepackage{makecell}
\usepackage{graphicx}
\usepackage{enumitem}
\usepackage{amsmath}
\usepackage{booktabs}
\usepackage{array}
\usepackage{graphicx}
\usepackage{multirow}
\usepackage{bbding}
\usepackage{colortbl}
\definecolor{mygray}{gray}{.92}
 
\usepackage{amsmath,amsthm,amssymb}

\usepackage{color}
\usepackage{float}
\usepackage[T1]{fontenc}
\usepackage{algorithm}
\usepackage{algpseudocode}
\usepackage[utf8]{inputenc}
\usepackage{microtype}
\usepackage{inconsolata}
\usepackage{graphicx}
\newcommand{\graybg}{\cellcolor{gray!10}}

\icmltitlerunning{Deep Tabular Research}

\begin{document}

\twocolumn[
  \icmltitle{Deep Tabular Research via Continual Experience-Driven Execution}

  \icmlsetsymbol{equal}{*}

  \begin{icmlauthorlist}
    \icmlauthor{Junnan Dong}{equal,comp}
    \icmlauthor{Chuang Zhou}{equal,comp}
    \icmlauthor{Zheng Yuan}{comp}
    \icmlauthor{Yifei Yu}{comp}
    \icmlauthor{Qiufeng Wang}{comp}\\
    \icmlauthor{Yinghui Li}{comp}
    \icmlauthor{Siyu An\dag}{comp}
    \icmlauthor{Di Yin}{comp}
    \icmlauthor{Xing Sun}{comp}
    \icmlauthor{Feiyue Huang\dag}{yyy}\\
    Ruijin Hospital, Shanghai Jiaotong University\\Tencent Youtu Lab
  \end{icmlauthorlist}

  \icmlaffiliation{yyy}{Ruijin Hospital, Shanghai Jiaotong University}
  \icmlaffiliation{comp}{Tencent Youtu Lab}

  \icmlcorrespondingauthor{Feiyue Huang}{Ruijin Hospital, Shanghai Jiaotong University}
  \icmlcorrespondingauthor{Siyu An}{Tencent Youtu Lab}
  \icmlkeywords{Tabular Reasoning, Complex Unstructured Table, Agentic Planning}

  \vskip 0.3in
]

\printAffiliationsAndNotice{}  
\begin{figure*}
    \centering
    \includegraphics[width=0.9\linewidth]{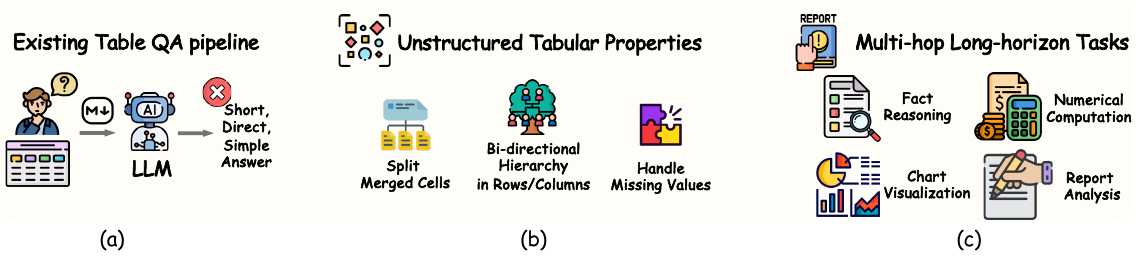}
    \caption{\small Existing Table QA pipelines (a) are limited to well-structured tables and shallow queries, and fail to handle unstructured tabular properties (b) and long-horizon analytical tasks (c), motivating our Deep Tabular Research.}
    \label{fig:running}
\end{figure*}

\begin{abstract}
Large language models often struggle with complex long-horizon analytical tasks over unstructured tables, which typically feature hierarchical and bidirectional headers and non-canonical layouts. We formalize this challenge as Deep Tabular Research (DTR), requiring multi-step reasoning over interdependent table regions. To address DTR, we propose a novel agentic framework that treats tabular reasoning as a closed-loop decision-making process. We carefully design a coupled query and table comprehension for path decision making and operational execution. Specifically, \((i)\) DTR first constructs a hierarchical meta graph to capture bidirectional semantics, mapping natural language queries into an operation-level search space; \((ii)\) To navigate this space, we introduce an expectation-aware selection policy that prioritizes high-utility execution paths; \((iii)\) Crucially, historical execution outcomes are synthesized into a siamese structured memory, i.e., parameterized updates and abstracted texts, enabling continual refinement. Extensive experiments on challenging unstructured tabular benchmarks verify the effectiveness and highlight the necessity of separating strategic planning from low-level execution for long-horizon tabular reasoning.
\end{abstract}

\section{Introduction}
Large language models (LLMs) have demonstrated remarkable capabilities in reasoning over structured data, leading to their widespread adoption in tabular question answering~\cite{gong2020tablegpt, zhao-etal-2024-tapera,katsis2022ait}. By serializing tables into textual formats, prior research has shown that LLMs can effectively resolve factual and numerical queries over well-structured schemas. These advances have positioned LLMs as promising general-purpose interfaces for interacting with tabular data~\cite{ren2025deep, somvanshi2024survey}. Despite this progress, most existing approaches rely on clean schemas, flat headers, and single-pass reasoning pipelines. Such assumptions severely limit their applicability to practical downstream scenarios where tables are frequently irregular, incomplete, and semantically implicit for comprehension.

Real-world tabular data, particularly spreadsheets, exhibit a wide range of unstructured properties that defy traditional TableQA pipelines. As illustrated in Figure~\ref{fig:running}, these tables often feature hierarchical and bi-directional headers, merged cells, and values that are missing or contextually defined. Navigating these complex structures remains a persistent challenge~\cite{tang2025st, wu2025realhitbench}. Furthermore, beyond structural complexity, practical analytical tasks are inherently long-horizon and multi-hop. A single query may necessitate a sequence of factual checks, numerical computations, and aggregations across disparate table regions. Answering such analytical queries requires more than simple retrieval; it demands iterative verification and conditional branching, where intermediate results must be scrutinized and revised before reaching a validated conclusion.

We formalize this problem as Deep Tabular Research (DTR), i.e., long-horizon complex tabular reasoning tasks that require coordinated data acquisition, computation, and analytical synthesis. While conventional approaches primarily rely on in-context learning, treating tables as text for direct LLM reasoning—such a paradigm is inherently limited by token constraints and struggles with precise numerical operations over large, irregular headers~\cite{sarkar2023testing,singha2023tabular}. To overcome these limitations, we advocate for a programmatic execution approach, leveraging tools like DataFrames to handle data processing and structural navigation. However, transitioning from static text-level reasoning to a code-driven agent introduces two significant challenges: \((i)\) Prohibitive search space for programmatic planning. Unlike static language reasoning for conventional table QA, translating high-level analytical intent (e.g., 'summarize by department') into concrete code operations (e.g., \texttt{df.groupby()}, \texttt{pd.pivot\_table()}) over unstructured tables involves a massive space of potential execution paths. Given the ambiguity of hierarchical headers and missing values, identifying the optimal sequence of operators is non-trivial. \((ii)\) Errors could inevitably propagate during complex long execution. Execution exposes concrete errors and ambiguities, while there is a limited mechanism to properly learn from past execution
outcomes, especially for failures, to guide future decisions. 

In this paper, we propose a novel agentic framework for DTR that treats tabular reasoning as a continual decision process driven by execution experience. Our framework explicitly decouples high-level strategic planning from low-level execution, ensuring that reasoning is informed by accumulated feedback rather than rigid heuristics. Specifically, our approach comprises three key components. \((i)\) A query-decomposed operator module maps natural language queries into a structured space of analytical operators, facilitating flexible composition; \((ii)\) We then design an expectation-aware policy that identifies promising execution trajectories under uncertainty, balancing exploration and utility without exhaustive simulation; \((iii)\) Finally, a siamese structured memory module records execution outcomes and failures, enabling the system to refine its planning strategy through experience. The memory is carefully evolved in a siamese mode inlucding both parameterized updates and abstracted textual experience. By grounding reasoning in verified micro operations and continuously adapting to feedback, our framework achieves robust error isolation and recovery across diverse tabular settings. Extensive experiments demonstrate that our approach consistently outperforms strong baselines on challenging unstructured benchmarks, establishing continual experience-driven execution as a superior foundation for deep tabular research.

\textbf{Contributions}:
\begin{itemize}[leftmargin=*] 
\item \textbf{Task Formalization:} We define the Deep Tabular Research (DTR) task, shifting the focus from simple TableQA to long-horizon, multi-hop analytical reasoning over unstructured, non-canonical tables. 
\item \textbf{Closed-Loop Agentic Framework:} We introduce a principled framework that decouples macro planning from micro execution, treating reasoning as an iterative decision-making process. 
\item \textbf{Experience-Driven Optimization:} We propose an expectation-aware selection mechanism and a structured memory graph that enable the agent to learn from execution feedback and mitigate error propagation. 
\item \textbf{Empirical Validation:} We provide extensive evaluations on unstructured tabular benchmarks, demonstrating the effectiveness and efficiency of DTR in handling complex, real-world data layouts. 
\end{itemize}

\section{Task Definition}
\begin{figure*}
    \centering
    \includegraphics[width=0.95\linewidth]{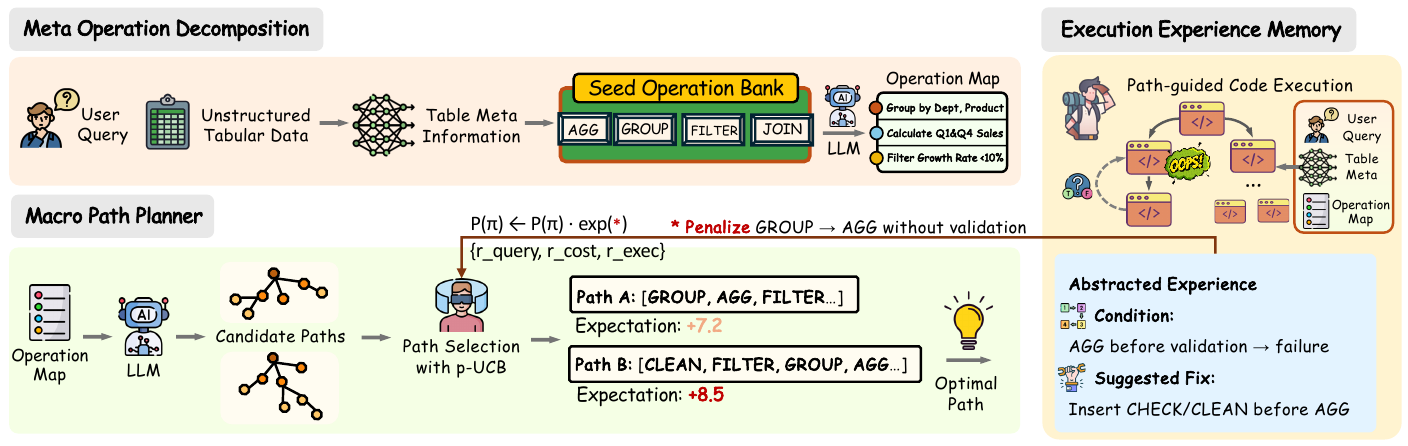}
    \caption{A sketched overview for our proposed Deep Tabular Research framework for complex unstructured tabular reasoning. DTR decomposes analytical intent into meta operations, plans executable macro paths via expectation-guided search, and executes them with an experience-aware memory that records structured execution feedback and updates planning policies across iterations.}
    \label{fig:figure1}
\end{figure*}
Deep Tabular Research (DTR) generalizes the problem of answering complex, multi-step analytical queries over non-canonical tables. Formally, a DTR task is defined by the tuple $(\mathcal{T}, \mathcal{Q}, \mathcal{E}, \mathcal{Y})$, representing the tabular domain, the query space, the execution environment, and the output space, respectively.

\fbox{\parbox{\linewidth}{
Let $T \in \mathcal{T}$ be an unstructured table with flexible schemas, $T$ may contain hierarchical headers, bidirectional row-column spans, and implicit semantic relations. Given a natural language query $q \in \mathcal{Q}$ that requires long-horizon reasoning (e.g., trend analysis, cross-region comparison, or recursive aggregation), the task is to produce a response $y \in \mathcal{Y}$ that is both factually accurate and computationally grounded.}}
\subsection{Execution-Grounded Paradigm}
Unlike traditional TableQA with a pure linguistic mapping $f(q, T) \to y$, DTR defines reasoning as an interactive exploration within an execution environment $\mathcal{E}$, where the agent incrementally executes operations and adapts its strategies.

\begin{itemize}[leftmargin=*]\item \textbf{Latent Structural State:} The true semantic structure of $T$ (e.g., the exact scope of a hierarchical header) is latent and must be inferred through interaction since raw data rarely presents explicit hierarchical dependencies.\item \textbf{Analytical Trajectory:} To resolve $q$, a sequence of atomic analytical actions $\mathbf{a} = (a_1, a_2, \dots, a_H)$ must be performed. Each $a_i$ represents a data-level transformation that maps an input data state to an intermediate result $o_i$.\item \textbf{State Transition:} The reasoning state at step $t$ is defined as $\mathcal{S}_t = \{q, T, (a_1, o_1), \dots, (a_{t-1}, o_{t-1})\}$. The objective is to find a trajectory of actions such that the final outcome $o_H$ supports the generation of correct output $y$.\end{itemize}



\vspace{-2mm}
\section{Deep Tabular Research}\label{sec:method}
DTR tackles long-horizon analytical reasoning over unstructured tables by formulating tabular reasoning as a closed-loop decision process. Figure~2 provides an overview of the framework. We demonstrate the core idea and technical details from five main aspects hereunder.
\vspace{-2mm}
\subsection{Tabular Comprehension and Structural Modeling}

Given a raw table $T$ (e.g., an Excel sheet) and a natural language query $q$, the first stage of DTR constructs a structured representation that captures both explicit and implicit table semantics. This representation serves as the foundation for downstream reasoning, enabling robust navigation of complex row--column relationships and hierarchical headers.
\vspace{-3mm}
\paragraph{Meta Information Extraction.}
Effective tabular reasoning requires a global understanding of table structure and salient semantic signals. We extract column and row headers, including sub-headers that encode hierarchical organization, as they provide concise summaries of how data is arranged. Beyond explicit headers, we identify implicit metadata such as measurement units, temporal or categorical markers, and aggregation indicators. Integrating both explicit and implicit signals yields a structured representation that captures the table’s overall organization and supports subsequent reasoning.
\vspace{-3mm}
\paragraph{Bi-directional Header Identification.}
Real-world tables often contain headers along both rows and columns, frequently with multi-level spans. We identify header regions along each axis and resolve their scopes via span alignment, producing a bidirectional header structure. Each data cell is associated with both row-wise and column-wise semantic descriptors, jointly defining its contextual meaning. These headers may form nested hierarchies, which are explicitly organized into a graph structure to capture interactions between row and column semantics.
\vspace{-5mm}
\paragraph{Meta Graph Construction.}
After extracting metadata from irregular table formats, the unstructured entities are organized into a structured graph denoted as $\mathcal{G}_T = (V_T, E_T)$. Each node in the graph corresponds to a header or content element, and edges represent containment or hierarchical relationships. Due to the bi-directional nature of table headers, the same sub-item can simultaneously belong to both row-wise and column-wise parent nodes, forming a overlapping tree-like hierarchical structure. This graph explicitly preserves the organizational layout of the table and serves as a structured representation for downstream reasoning tasks.
\vspace{-7mm}

\subsection{Query-Guided Operation Mapping}
\vspace{-2mm}
Given a natural language query $q$ and the constructed table graph $\mathcal{G}_T$, we next determine a sequence of operations that can be executed over the graph to answer the query. Rather than directly reasoning over raw table cells, our approach relies on a predefined seed operation bank that encapsulates a diverse set of atomic operations commonly required for tabular reasoning. The seed operation bank contains a collection of basic operations such as filtering, group aggregation, and numerical sorting: $ \mathcal{O} = \{\texttt{CLEAN}, \texttt{FILTER}, \texttt{GROUP}, \texttt{AGG}, \texttt{JOIN}, \texttt{SORT}, \dots\}.
$


To align the query with executable operations, we leverage an LLM-based agent to perform decision making over both the decomposed sub-queries and the structured table graph. The graph $\mathcal{G}_T$ is linearized and provided to the agent in the form of relational triples, such as
(\textit{table}, \textit{has\_column\_header}, \textit{column\_header\_description}) and
(\textit{column\_header}, \textit{has\_child}, \textit{child\_header}),
which explicitly encode the hierarchical and containment relations within the table. Given each sub-query and the corresponding graph description, the agent selects a set of candidate operations from the seed operation bank that are most relevant to the reasoning intent. The selected operations for query $q$ are denoted as $\mathcal{O}(q) = \{ o_1, o_2, \dots, o_K \}$, forming the basis for subsequent execution and reasoning over the table graph. The following text box provides a concrete input example. 

\paragraph{Operation Map Construction.}
We further construct an operation map that encodes dependencies and admissible orderings among operations. Rather than treating candidate actions as an unordered set, the operation map organizes them into a sequential path that respects logical and semantic constraints. Certain operations impose prerequisite contexts; for instance, \texttt{AGG} requires a well-defined grouping scope, while \texttt{FILTER} may be applied either before or after aggregation depending on whether it constrains raw values or aggregated results. These initial operation paths are subsequently refined through more sophisticated selection mechanisms to identify the most coherent sequence.

\subsection{Path Planning with Expectation-Aware Selection}
Given the feasible operation paths constructed from the operation map, a path is defined as $\pi = (o_1, o_2, \dots, o_L)$, representing an ordered sequence of operations. DTR performs path planning to identify the most promising execution strategy for a given query. Rather than treating all admissible paths equally, we introduce an expectation-aware selection that evaluates candidate paths based on their anticipated utility. We enumerate a finite set of candidate paths $\{\pi_i\}$ by composing operations under dependency constraints and pruning invalid or redundant sequences. Importantly, path selection is not performed in a single pass. As operations are executed, DTR reflects on the intermediate execution results and iteratively updates its preference over candidate paths. This closed-loop refinement allows the planner to revise earlier decisions, and progressively focus on operation sequences that are more likely to yield correct outcomes.

\paragraph{Expectation-Aware Scoring.}

For each candidate path $\pi$, we maintain a set of path-level statistics that summarize its historical execution behavior, including an estimate of its expected return $\hat{R}(\pi)$, the number of times the path has been executed $N(\pi)$, and a prior term $P(\pi)$ that reflects structural plausibility or domain knowledge. Based on these terms, we define an expectation-aware score for each path:
\begin{equation}\small
\mathcal{E}(\pi)
=
\hat{R}(\pi)
+
\alpha \cdot P(\pi)
\sqrt{
\frac{\log \sum_{\pi'} N(\pi')}{1 + N(\pi)}
}.
\end{equation}
The first term encourages exploitation by favoring paths that have produced reliable intermediate or final results in previous executions. The second term promotes exploration by assigning higher scores to paths that are structurally plausible but have been executed fewer times, with the logarithmic normalization accounting for the overall execution budget. The hyperparameter $\alpha$ controls the trade-off between exploitation and exploration. Our framework selects and executes the path that maximizes the expectation-aware score $\mathcal{E}(\pi)$, enabling a balance between enhancing reasoning strategies and discovering alternative paths.

\paragraph{Theoretical Boundedness.}
Assume the realized execution reward is bounded such that $R(\pi) \in [0, R_{\max}]$, and the structural prior satisfies $P(\pi) \in [0,1]$. Since $\hat{R}(\pi)$ is an empirical estimate of $R(\pi)$, it follows that
$\hat{R}(\pi) \le R_{\max}$. Therefore, the expectation score is upper bounded by:
\begin{equation}\small
\mathcal{E}(\pi)
\le
R_{\max}
+
\alpha \sqrt{\log \sum_{\pi'} N(\pi')}.
\vspace{-1mm}
\end{equation}
This bound ensures that expectation values remain scaled and prevents unbounded optimism during exploration. Moreover, the exploration term for a fixed global execution budget is monotonically decreasing with respect to $N(\pi)$:
\begin{equation}\small
\lim_{N(\pi) \to \infty}
\alpha \cdot P(\pi)
\sqrt{
\frac{\log \sum_{\pi'} N(\pi')}{1 + N(\pi)}
}
= 0.
\end{equation}
As a result, paths that are sufficiently explored gradually shift from exploration-driven selection to exploitation based on empirical performance obtained from current executions.



\paragraph{Path Selection.}

The planner selects the top-$k$ paths with the highest expectation scores for execution. Each selected path is then instantiated into an executable analytical program and applied to the table, producing intermediate or final results. Importantly, expectations are evaluated at the path level, since rewards in long-horizon tabular reasoning are inherently not decomposable and cannot be attributed to individual intermediate operations. 
Execution outcomes are evaluated against the query intent to obtain a scalar feedback signal, which is subsequently used to update the path-level statistics. Through this iterative process, DTR progressively refines its preference over operation paths, enabling more accurate and reliable planning for complex tabular tasks.

\paragraph{Iterative Interaction.} During the execution of a selected operation path, DTR enables intermediate interactions with the LLM agent between consecutive operations. Specifically, before and after executing each operation, the agent is prompted to produce a discrete flag signal that characterizes the latest execution stage, i.d., [THINK] / [CODE]. The flag indicates whether the agent is engaged in \emph{coding}, i.e. running executable code, or in \emph{thinking}, such as summarizing intermediate results and validating the ongoing analytical logic. Depending on the flag type, the corresponding execution code or reasoning summary is recorded as part of the execution trace. Prior to executing the next operation, this context is fed back to the agent, providing an explicit description of the current execution state and reasoning trajectory. This design allows the agent to adapt subsequent decisions based on both concrete execution results and evolving analytical understanding, facilitating coherent progression along the operation path.

\subsection{Siamese Experience-Guided Reflection}

Beyond path-level planning, DTR further incorporates a siamese experience-guided execution mechanism that leverages feedback at two complementary levels to perform cross-path reasoning and consolidation. The first type is \emph{parameterized execution feedback}, which captures concrete execution signals produced by running a selected operation path on the table. These signals include execution success or failure, and intermediate result validity. Such feedback provides fine-grained and specific supervision that reflects how well a particular operation performs on the given table.

In parallel, DTR maintains an \emph{abstracted experience} channel that summarizes execution outcomes accumulated up to each timestamp. Instead of retaining raw execution traces, this channel distills higher-level patterns, such as which classes of operation paths tend to be effective under certain query structures or table organizations. These abstracted experiences are agnostic to specific table values, enabling transfer across instances and supporting more robust decision making in future executions. These two streams operate in a siamese manner: parameterized execution feedback informs immediate path refinement for the current query, while abstracted experience guides longer-term preference. Together, they form a closed-loop execution framework.

\paragraph{Parameterized Execution Feedback Signals.}
Given a selected operation path $\pi = (o_1, \dots, o_L)$ and its instantiated execution program $\mathcal{P}_\pi$, DTR executes $\mathcal{P}_\pi$ on the table and collects parameterized execution feedback signals that reflect the observed execution behavior. We denote the execution feedback as the following expression:
\begin{equation}\small
\label{f_pi}
\mathbf{f}(\pi) = \big( f_{\text{exec}}(\pi),\; f_{\text{time}}(\pi),\; 
f_{\text{type}}(\pi) \big),
\end{equation}
where each component evaluates a different aspect of execution behavior. The execution validity signal $f_{\text{exec}}(\pi) \in \{0,1\}$ indicates whether the program runs without errors. The execution time signal $f_{\text{time}}(\pi) \in \mathbb{R}^+$ records the time required to complete the execution of $\mathcal{P}_\pi$, reflecting the computational efficiency of the operation path. The type consistency signal metric $f_{\text{type}}(\pi) \in \{0,1\}$ evaluates whether the intermediate and final results match the expected output format implied by the query and operation sequence, such as whether a ranking operation yields an ordered list, or a numerical aggregation produces a scalar value. These signals are parameterized by the operation sequence, allowing DTR to distinguish between structurally similar paths that differ in execution behavior. The overall execution reward is computed as $r(\pi) = \phi\big( \mathbf{f}(\pi) \big)$, which is used to update path-level expectations and guide subsequent decisions.


\paragraph{Abstracted Experience Feedback.} In addition to quantitative evaluation, abstracted experience is used to guide subsequent path execution. Rather than encoding concrete signals such as running time or output validity, this feedback summarizes semantic and strategic observations distilled from executed actions, preserving flexibility. For example, recurring aggregation failures may trigger the insertion of validation or cleaning operations prior to aggregation. 
\begin{tcolorbox}[
    colback = blue!5!white, colframe = blue!75!black, arc = 3pt, boxrule = 1pt
]
{  \footnotesize
\textbf{Query: }What are the top-selling categories in Q3? \\ 
\textbf{Feedback Summary: } \\
- Path \(\pi_1\) first grouped seasons to compute total sales, then filtered Q3; filtering operation should be executed first to avoid computation of irrelevant data. \\  
- Error: \textit{category} not found. Action: re-read table samples and derive sub-column \textit{product category}.\\ 
- Repeated sorting operations before and after grouping do not contribute and can be skipped.  \\
\textbf{Execution Reflection:} maintaining the correct filter-then-aggregate order; reading the sub-column \textit{product category}; sorting operation placed at the end.
}

\end{tcolorbox}

\begin{table*}[ht!]\footnotesize
\centering
\caption{Full Comparisons over DTR-Bench to evaluate over accuracy, quality, and efficiency.}
\resizebox{0.98\textwidth}{!}{%
\begin{tabular}{@{}l|cc|cc|cc|cc|ccc@{}}
\toprule
\multirow{4}{*}{\textbf{Methods}} & \multicolumn{2}{c|}{\textbf{Accuracy}} & \multicolumn{2}{c|}{\textbf{Analysis Depth}} & \multicolumn{2}{c|}{\textbf{Feasibility}} & \multicolumn{2}{c|}{\textbf{Aesthetics}} & \multirow{3}{*}{\makecell{\textbf{Avg.}\\\textbf{Runtime}\\\textbf{(s)}$\downarrow$}} & \multirow{3}{*}{\makecell{\textbf{Total}\\\textbf{Output}\\\textbf{Tokens}$\downarrow$}} & \multirow{3}{*}{\makecell{\textbf{Avg.}\\\textbf{LLM}\\\textbf{Calls}}$\downarrow$} \\
\cmidrule(lr){2-3} \cmidrule(lr){4-5} \cmidrule(lr){6-7} \cmidrule(lr){8-9}
& \makecell{Win Rate} & \graybg\makecell{Score Rate} & \makecell{Win Rate} & \graybg\makecell{Score Rate} & \makecell{Win Rate} & \graybg\makecell{Score Rate} & \makecell{Win Rate} & \graybg\makecell{Score Rate} & & & \\
& {(Tie=0)$\uparrow$} & \graybg{ (Tie=0.5)$\uparrow$} & { (Tie=0)$\uparrow$} & \graybg{(Tie=0.5)$\uparrow$} & {(Tie=0)}$\uparrow$ & \graybg{(Tie=0.5)$\uparrow$} & { (Tie=0)$\uparrow$} & \graybg{(Tie=0.5)$\uparrow$} & & & \\
\midrule
\multicolumn{12}{c}{\textbf{Table specific LLM}} \\
\midrule
TableGPT2-7B & 0.20 & \graybg 8.41 & 5.12 & \graybg 5.12 & 4.33 & \graybg 4.35 & 6.21 & \graybg 6.21 & 3.42 & 12,450 & 1.0 \\
TableLLM-7B & 0.15 & \graybg 6.22 & 3.84 & \graybg 3.84 & 3.10 & \graybg 3.12 & 4.55 & \graybg 4.55 & 3.15 & 10,820 & 1.0 \\
StructGPT & 0.10 & \graybg 4.15 & 2.10 & \graybg 2.10 & 1.85 & \graybg 1.86 & 2.30 & \graybg 2.30 & 4.88 & 8,420 & 1.0 \\
\midrule
\multicolumn{12}{c}{\textbf{Common LLM}} \\
\midrule
DeepSeek-V3 & 1.21 & \graybg 30.2 & 21.72 & \graybg 21.72 & 20.60 & \graybg 20.60 & 31.23 & \graybg 31.23 & 38.64 & 49,271,1 & 1.0 \\
DeepSeek-V3.2 & 1.28 & \graybg 33.52 & 25.22 & \graybg 25.22 & 24.42 & \graybg 24.43 & 36.63 & \graybg 36.83 & 39.02 & 52,446,3 & 1.0 \\
\midrule
\multicolumn{12}{c}{\textbf{Workflow}} \\
\midrule
ST-Raptor & 0.62 & \graybg 22.40 & 6.00 & \graybg 6.00 & 7.41 & \graybg 7.41 & 12.40 & \graybg 12.40 & 999.16 & 31,077,2 & 9.2 \\
TreeThinker & 1.83 & \graybg 31.00 & 22.82 & \graybg 22.82 & 21.42 & \graybg 21.43 & 36.83 & \graybg 36.83 & 140.78 & 17,067,00 & 5.1 \\
Code Loop & 1.32 & \graybg 27.50 & 9.40 & \graybg 9.51 & 14.81 & \graybg 14.92 & 20.42 & \graybg 20.42 & 175.84 & 75,204,2 & 8.8 \\
\midrule
\multicolumn{12}{c}{\textbf{DTR}} \\
\midrule
DTR (DS-v3) & \textbf{1.93} & \graybg \textbf{37.53} & \textbf{30.23} & \graybg \textbf{30.23} & \textbf{27.62} & \graybg \textbf{27.64} & \textbf{42.74} & \graybg \textbf{42.64} & 62.09 & 81,754,2 & 4.7 \\
\bottomrule
\end{tabular}%
}
\end{table*}

\subsection{Reflection-Driven Path Adaption}

With real-time experience-guided reflection from the siamese mechanism, candidate operation paths are dynamically updated, leading to corresponding changes in their expectation scores. This allows DTR to progressively favor paths that demonstrate consistent effectiveness, while enabling subsequent operation selection to learn from accumulated reflections.

\paragraph{Continual Expectation Update.}
After executing an operation path $\pi$, DTR aggregates execution outcomes into a path-level reward $R(\pi)$, reflecting the overall effectiveness of the reasoning trajectory. The estimated expected return is updated incrementally as
\begin{equation}\small
\hat{R}(\pi) \leftarrow (1-\eta)\,\hat{R}(\pi) + \eta \cdot R(\pi),
\vspace{-1mm}
\end{equation}
where $\eta \in (0,1]$ is a learning rate controlling the influence of newly observed execution outcomes. While $R(\pi)$ is only defined for executed paths, $\hat{R}(\pi)$ is continuously updated for paths to be executed. Through this mechanism, feedback from one path could influence the expectations of other related paths. As defined in Equation \ref{f_pi}, the real record of execution-level factors such as running time and output format consistency is reflected in the reward $r(\pi)$. Besides, abstracted experience is utilized to guide adjustments in operation selection for other candidate paths, thereby modifying their expected returns. For instance, structural modifications including operation insertions or removals lead to corresponding changes in execution count $N(\pi)$. This allows our framework to dynamically prioritize paths that are structurally aligned with previously successful signals. 
\vspace{-4mm}
\paragraph{Closed-Loop Optimization.}
DTR performs closed-loop optimization by path planning, execution, and expectation update in a progressive manner. At each iteration $t$, the agent selects the candidate path with the highest expectation scores under real-time estimate $\mathcal{E}_t(\pi)$. These paths are executed to produce intermediate or final results, whose execution feedback is then used to update both parameterized signals and abstracted experience. To determine the final answer, DTR aggregates outputs from multiple executed paths.
Let $\mathcal{A} = \{a_1, \dots, a_m\}$ denote the set of candidate answers and the final answer $a^\ast$ is selected by majority agreement:
\vspace{-2mm}
\begin{equation}\small
a^\ast = \arg\max_{a \in \mathcal{A}} \sum_{i=1}^{m} \mathbb{I}(a_i = a),
\vspace{-2mm}
\end{equation}
where only answers satisfying the query requirements and correct formats are considered. This voting-based selection improves robustness against individual execution errors and reinforces the reliability supported by multiple trials.
\section{Experimental Analysis}
\label{exp: dtr}

\begin{table*}[ht!]\scriptsize
\centering
\caption{Comparisons over the RealHitBench dataset for five different task types.}
\resizebox{0.98\textwidth}{!}{%
\begin{tabular}{@{}l|cc|cc|cc|cc|ccc@{}}
\toprule
\multirow{3}{*}{\textbf{Methods}}

& \multicolumn{2}{c}{\makecell{\textbf{Fact}\\\textbf{Checking}}} & \multicolumn{2}{c}{\makecell{\textbf{Numerical}\\\textbf{Reasoning}}} & \multicolumn{2}{c}{\makecell{\textbf{Structure}\\\textbf{Comprehending}}} & \multicolumn{2}{c}{\makecell{\textbf{Data}\\\textbf{Analysis}}} & \multicolumn{2}{c}{\makecell{\textbf{Chart/Report}\\\textbf{Generation}}} \\
\cmidrule(lr){2-3} \cmidrule(lr){4-5} \cmidrule(lr){6-7} \cmidrule(lr){8-9} \cmidrule(lr){10-11}
& EM$\uparrow$ & F1$\uparrow$ & EM$\uparrow$ & F1$\uparrow$ & EM$\uparrow$ & F1$\uparrow$ & \makecell{LLM-EVAL}$\uparrow$ & ROUGE$\uparrow$ & \makecell{PASS@1}$\uparrow$ & ECR$\uparrow$ \\
\midrule
TableGPT2-7B & 46.10 & 53.80 & 29.31 & 39.81 & 48.23 & 56.68 & 62.76 & 33.25 & 32.47 & 67.53 \\
TableLLM-7B & 38.25 & 44.12 & 22.40 & 31.65 & 41.10 & 49.34 & 55.80 & 28.42 & 18.20 & 42.15 \\
StructGPT & 25.40 & 32.15 & 14.55 & 20.80 & 30.25 & 38.60 & 42.33 & 19.50 & 5.12 & 12.44 \\
\midrule
GPT4o & 43.39 & 51.87 & 27.63 & 36.68 & 42.68 & 52.89 & 65.24 & 33.10 & 10.39 & 25.32 \\
DeepSeek-v3 & 57.21 & 53.42 & 47.05 & 50.61 & 43.31 & 74.63 & 61.40 & 34.76 & 9.09 & 24.68 \\
\midrule
Code Loop (DeepSeek-v3) & 48.19 & 56.49 & 42.93 & 49.68 & 44.19 & 51.95 & 62.51 & 33.73 & 20.78 & 39.60 \\
Code Loop (Qwen3-1.7B) & 6.91 & 7.65 & 4.02 & 5.75 & 49.24 & 53.08 & 17.97 & 17.29 & 0.00 & 0.60 \\
Code Loop (Qwen3-4B) & 16.61 & 19.76 & 10.25 & 13.00 & 32.85 & 30.30 & 25.83 & 18.17 & 3.25 & 14.30 \\

\midrule
\midrule
DTR (Qwen3-1.7B) & 17.93 & 23.50 & 13.75 & 19.01 & 17.17 & 27.88 & 40.28 & 16.44 & 5.16 & 21.94 \\
DTR (Qwen3-4B) & 30.42 & 37.44 & 22.05 & 30.13 & 32.6 & 43.32 & 53.88 & 20.15 & 8.39 & 30.32 \\
DTR (DeepSeek-v3) &\textbf{58.22} &\textbf{64.47} &\textbf{55.51} &\textbf{61.98} &\textbf{56.57} &\textbf{77.95} &\textbf{70.90} &\textbf{38.67} &\textbf{52.69} & \textbf{100.00} \\							
\bottomrule
\end{tabular}%
\label{exp: real}
}
\end{table*}

We evaluate Deep Tabular Research (DTR) on a diverse set of unstructured tabular reasoning benchmarks, covering factual lookup, numerical computation, structural understanding, long-horizon data analysis, and visualization tasks. 
Our evaluation focuses on both \emph{task accuracy} and \emph{end-to-end analytical effectiveness}, reflecting the real-world demands of complex spreadsheet-based workflows. The details of the adopted benchmarks, state-of-the-art baselines and evaluation protocols are introduced in Appendix~\ref{implement}.

\subsection{Main Results}
Tables~\ref{exp: dtr} and Table~\ref{exp: real} report the main experimental results on DTR-Bench and RealHitBench, respectively. Though both benchmarks target table-centric reasoning, they emphasize fundamentally different aspects of model capability. For fair comparison, we allow non-code mode for the RealHitBench dataset since it focuses on answer-level correctness with concise formats. On DTR-Bench, DTR achieves the strongest overall performance across all dimensions, including accuracy, analysis depth, feasibility, and aesthetics. Notably, the improvements are consistent under both strict win-rate and more tolerant score-rate evaluation, indicating that DTR does not merely outperform baselines in marginal cases, but produces systematically higher-quality outputs.

These gains cannot be attributed solely to stronger backbone models. Compared to pure LLM baselines, DTR demonstrates substantially deeper analysis and higher feasibility. This suggests that unconstrained language generation, even in large models, has difficulty reliably organizing multi-step table operations into executable workflows. Conversely, agent-based frameworks such as ST-Raptor, TreeThinker, and Code Loop exhibit improved structural reasoning but incur significant computational overhead and instability due to extensive branching and repeated trial executions. DTR stands out in this trade-off space. By explicitly planning over macro-level operation paths and selecting candidates based on learned expectations derived from historical experience, DTR avoids exhaustive search while still preserving long-range reasoning coherence. As a result, it produces more complete and visually coherent analytical reports than pure LLMs, while remaining substantially more efficient than tree-based or loop-based agent frameworks.

Beyond quality, DTR also demonstrates favorable efficiency–performance balance. While workflow-based baselines often require an order of magnitude more LLM calls and runtime to achieve competitive analytical depth, DTR attains superior results with fewer calls and more predictable execution cost. This suggests that reasoning at the level of operation sequences, rather than token-level or step-level exploration, leads to more stable and scalable analytical behavior for complex table reasoning tasks.

\begin{table}[ht!]
\centering
\caption{Architecture ablation study of each components in DTR.}
\label{tab:architecture_ablation}
\resizebox{\linewidth}{!}{%
\begin{tabular}{lcccccccc}
\toprule
\textbf{Configuration} & \textbf{Meta} & \textbf{QDO} & \textbf{Exp.} & \textbf{Abst.} & \textbf{Acc.} & \textbf{Anal.} & \textbf{Feas.} & \textbf{Aesth.} \\
\midrule
+ Meta Info & \checkmark & $\times$ & $\times$ & $\times$ & 34.8 & 27.1 & 25.6 & 38.2 \\
+ QDO & \checkmark & \checkmark & $\times$ & $\times$ & 36.2 & 28.8 & 26.7 & 40.5 \\
+ Expectation & \checkmark & \checkmark & \checkmark & $\times$ & 37.1 & 29.6 & 27.2 & 41.8 \\
\textbf{DTR Full} & \checkmark & \checkmark & \checkmark & \checkmark & \textbf{37.5} & \textbf{30.2} & \textbf{27.6} & \textbf{42.7} \\
\bottomrule
\end{tabular}%
}
\vspace{-3mm}
\end{table}

\subsection{Ablation Studies}
\label{sec:ablation}
\subsubsection{Component Contributions}
We conduct a systematic ablation study to quantify the marginal contribution of each constituent module within our framework. Beginning with a \textbf{pure LLM baseline} (DeepSeek-V3), we progressively integrate \((i)\) tabular meta information, \((ii)\) query-to-operation decomposition, \((iii)\) expectation-aware macro path selection informed by historical execution feedback, and \((iv)\) abstracted execution experience. Table~\ref{tab:architecture_ablation} summarizes the results across multiple evaluation dimensions, including Accuracy, Analysis Depth, Feasibility, Aesthetics, and the mean number of LLM calls.

The results demonstrate that DTR yields consistent performance gains over the baseline, achieving a total accuracy improvement of \textbf{4.0} percentage points (33.5\% $\rightarrow$ 37.5\%). Among the individual components, tabular meta-information and query decomposition provide the most significant uplift (+1.3 and +1.4 points, respectively), which suggests that explicit structural grounding and operation-level intent modeling are fundamental to unstructured tabular reasoning. The incorporation of historical feedback for macro path selection further enhances accuracy by 0.9 points, indicating that high-level planning benefits substantially from past experience even in the absence of low-level code simulations. Finally, the inclusion of abstracted execution experience provides an additional 0.4 point gain, validating the utility of distilling raw execution outcomes into reusable knowledge for novel problem instances.


\begin{table}[ht!]
\centering
\caption{Exploration over prompting strategy for think and code.}
\label{tab:prompt_ablation}
\resizebox{\linewidth}{!}{%
\begin{tabular}{lcccc}
\toprule
\textbf{Prompt Strategy} & \textbf{Accuracy} & \textbf{Analysis} & \textbf{Code Error} & \textbf{Avg} \\
& & \textbf{Depth} & \textbf{Rate} & \textbf{Calls} \\
\midrule
No [THINK] (Direct) & 35.2 & 27.8 & 42.3\% & 5.8 \\
Simple [THINK] Hint & 36.4 & 28.9 & 35.6\% & 5.2\\
\textbf{[THINK]+[CODE]} & \textbf{37.5} & \textbf{30.2} & \textbf{28.4\%} & \textbf{4.78}\\
Multi-stage Reflection & 37.2 & 29.8 & 26.1\% & 5.5 \\
\bottomrule
\end{tabular}%
}
\end{table}
\subsubsection{Prompting Strategy Analysis}
\label{sec:prompt-ablation}

We observe diminishing returns as the system complexity increases. While each module contributes positive gains, the incremental improvements become smaller with additional components. This trend aligns with our objective of decoupling foundational capabilities, such as table comprehension and decomposition, from iterative refinement processes like experience-aware planning and memory abstraction.Given DTR's step-wise paradigm, code reliability is critical to both correctness and computational efficiency. We therefore evaluate various strategies during the execution phase. Table~\ref{tab:prompt_ablation} compares four configurations, including direct code generation without explicit reasoning steps, a lightweight reasoning hint, our default structured [THINK]+[CODE] prompting mode, and a multi-stage reflection strategy.

The structured [THINK]+[CODE] scheme well balances between performance and efficiency. It attains the highest accuracy (37.5\%) and analysis depth (30.2) while simultaneously reducing the code error rate from 42.3\% to 28.4\% compared to the direct generation baseline. Furthermore, this approach optimizes the average number of LLM calls (4.78 versus 5.8), which implies that isolating semantic reasoning from code emission enhances execution stability and mitigates redundant retry cycles. Although multi-stage reflection achieves a slightly lower error rate (26.1\%), it introduces substantial runtime overhead (48.6s versus 42.1s), suggesting that additional reflection yield marginal benefits.
\vspace{-3mm}
\subsection{Efficiency and Scalability}
\vspace{-1mm}
\label{sec:efficiency}
\begin{figure}
    \centering
    \includegraphics[width=0.85\linewidth]{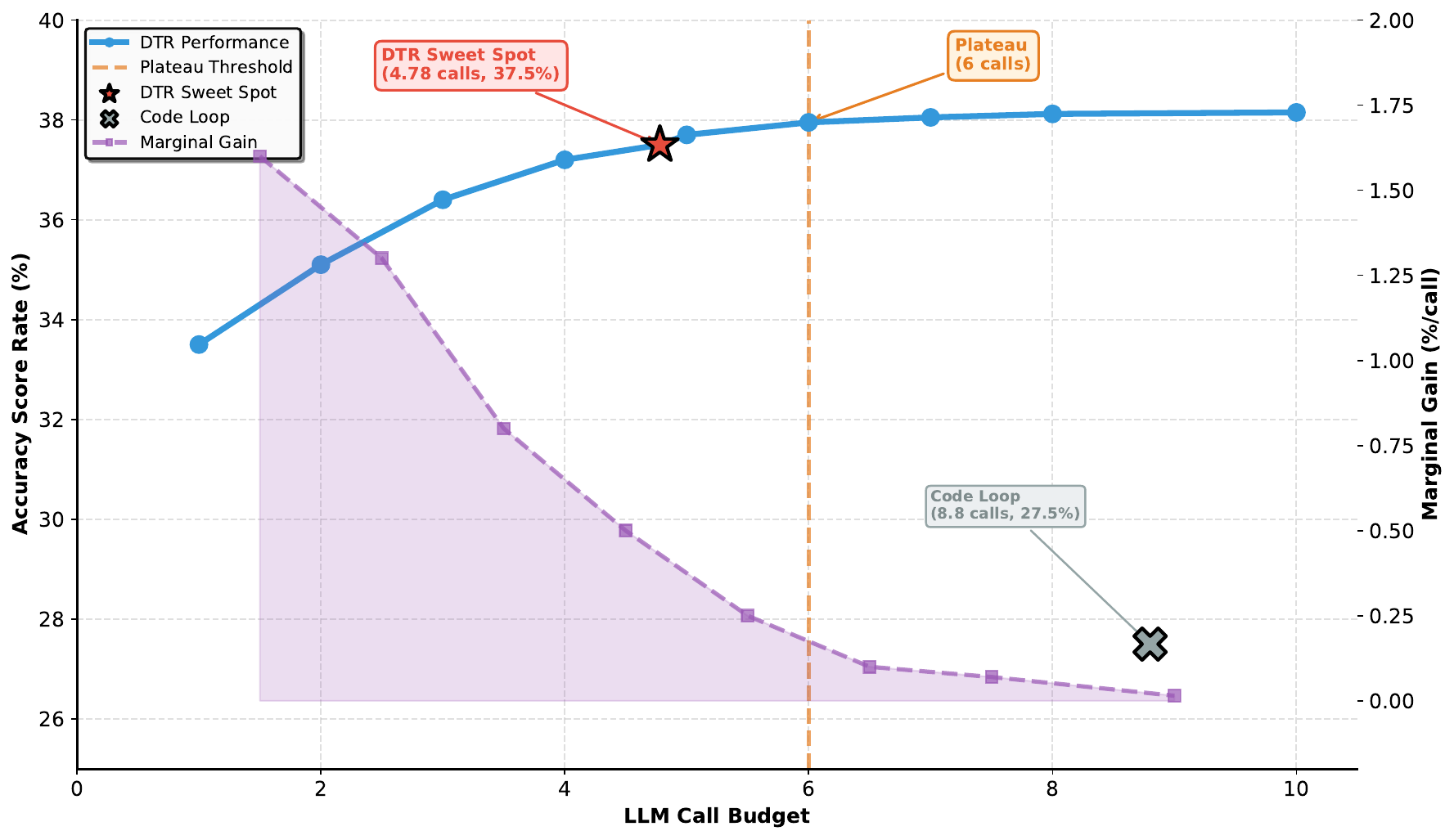}
    \caption{\footnotesize LLM call budget analysis. The blue curve shows performance (left y-axis) and the purple curve shows marginal gain (right y-axis). DTR's 4.78-call configuration red star) achieves optimal efficiency by avoiding the plateau region.}
    \label{fig:budget}
\vspace{-6mm}
\end{figure}

We assess the computational efficiency of DTR by comparing model performance and the frequency of LLM calls, using the call budget as a proxy for inference-time cost. Figure~\ref{fig:budget} presents the performance trajectory as the budget increases and the marginal utility per additional call.

The analysis reveals three distinct regimes. In the rapid growth phase (1--3 calls), performance scales sharply with high marginal gains averaging approximately +1.45\% per call, demonstrating that even sparse iterative interactions significantly bolster long-horizon tabular reasoning. In the transitional regime (3--6 calls), improvements persist but decelerate to approximately +0.45\% per call as the system nears its performance ceiling. Beyond 6 calls, each additional call yields less than 0.15\% performance gain. DTR operates at an average of \textbf{4.78} calls, effectively positioning it within the optimal transition region where quality and computation are balanced. In contrast, the CodeLoop baseline exhibits a failure mode characterized by over-iteration. Despite exhausting a significantly higher budget (8.8 calls), it achieves only 27.5\% accuracy, suggesting that unconstrained execution without a strategic selection mechanism can propagate errors and degrade overall performance. These findings justify the importance of DTR’s budgeted design and its expectation-aware path selection.


\vspace{-1mm}
\subsection{Case Study: Planning Dynamics}
\vspace{-1mm}
\label{sec:case-study}
\begin{figure}
    \centering
    \includegraphics[width=0.92\linewidth]{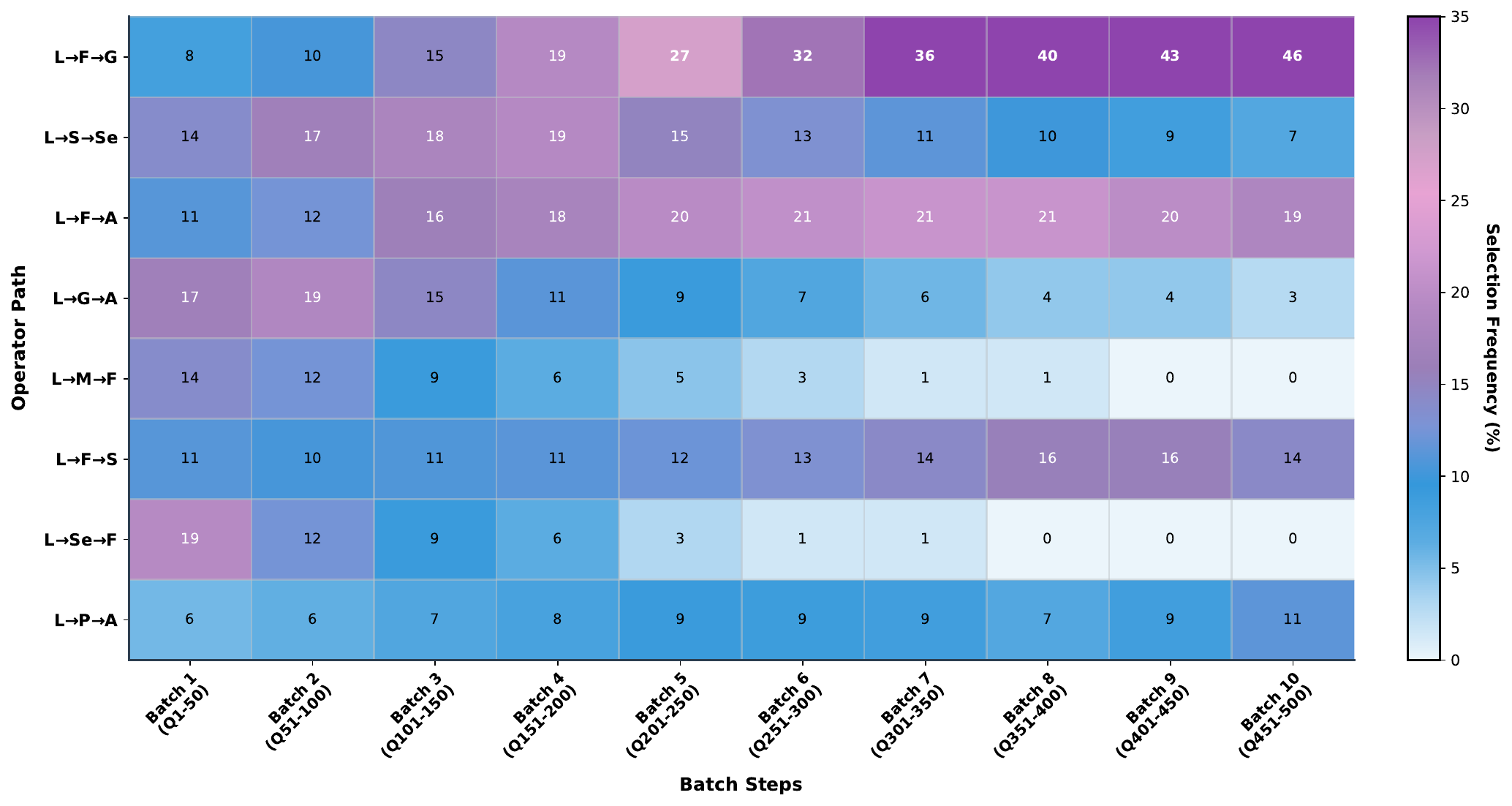}
    \caption{\footnotesize Path selection evolution across 10 batches. Colors changing from light blue to deep purple indicate  both exploration and exploitation, respectively.}
    \label{fig:path-evolution}
    \vspace{-5mm}
\end{figure}

To further investigate the internal dynamics of DTR, we visualize the evolution of macro path selection over 500 queries partitioned into 10 sequential batches. Figure~\ref{fig:path-evolution} depicts the selection frequency across eight operator paths via a heatmap, where increased density represents higher preference. During the initial batches (1--3), the system engages in broad exploration by selecting candidate paths with near-uniform frequency (3--7\% each). This behavior reflects high initial uncertainty and the necessity of empirical evaluation under diverse tabular conditions.  There exhibits a clear convergence pattern by the fifth batch. Path 0 (\texttt{LOAD} $\rightarrow$ \texttt{FILTER} $\rightarrow$ \texttt{GROUPBY}) increases in frequency from 3\% to 28\%, while paths with consistently low rewards are automatically pruned, e.g., Path 7 which declines to near zero.

Notably, DTR avoids collapsing into a single deterministic strategy. In the final batches (8--10), the primary path stabilizes at roughly 31\% while Path 5 remains a robust secondary option at 11\%. The remaining probability mass is distributed across various alternative paths to maintain approximately 10--15\% exploration. This equilibrium indicates that DTR successfully adapts through execution experience by prioritizing high-return strategies while retaining sufficient diversity for context-sensitive reasoning, demonstrating a balanced exploitation-exploration trade-off.
\vspace{-2mm}
\section{Conclusions}
\vspace{-2mm}
In this paper, we formally define \emph{Deep Tabular Research} as a new paradigm of long-horizon analytical reasoning tasks over unstructured tables. We propose a principled agentic framework, i.e., DTR, that treats tabular reasoning as a closed-loop decision-making process grounded in executable operations. DTR jointly optimizes strategic planning and operational execution through query-aware operator abstraction, expectation-driven path selection, and experience-based memory refinement, enabling robust reasoning under structural ambiguity and execution uncertainty. Extensive experiments on challenging unstructured tabular benchmarks demonstrate consistent improvements over SOTA baselines in reasoning accuracy, execution stability, and efficiency. These results highlight the necessity of separating high-level planning from low-level execution and establish execution-driven, experience-aware reasoning as a foundational paradigm for deep tabular research.

\section*{Broader Impact}

We present a framework for advancing machine learning systems in complex tabular reasoning and analytical tasks for social good based on large language models. Our primary goal is to improve the robustness, interpretability, and effectiveness of automated reasoning over structured data, which is a core and well-established research direction in machine learning. Potential positive societal impacts of this work include improved automation and decision support in data-driven domains such as scientific analysis, business intelligence, and public data reporting, where accurate and transparent table-based reasoning is essential. By enabling models to better understand complex table structures and perform multi-step analytical reasoning, we may help reduce manual effort and errors in data analysis workflows.

At the same time, we do not anticipate significant new ethical risks beyond those commonly associated with large language models and automated data analysis systems. As with prior work in this area, misuse could arise if such systems are deployed without appropriate human oversight, particularly in high-stakes settings involving sensitive or biased data. These risks are not unique to our method and are best addressed through established practices such as responsible deployment, data governance, and human-in-the-loop validation. We believe this work contributes incrementally and responsibly to the broader field of machine learning, without introducing novel ethical concerns that would require special mitigation beyond existing standards.

\newpage
\bibliography{example_paper}
\bibliographystyle{icml2026}

\newpage
\appendix
\onecolumn
\appendix
\section{DTR Algorithm}
\begin{algorithm}[ht!]
\caption{Deep Tabular Research}
\label{alg:dtr-framework}
\begin{algorithmic}[1]
\Require Query $q \in \mathcal{Q}$, Table $T \in \mathcal{T}$, Execution Environment $\mathcal{E}$, Learning Rate $\alpha$, Number of candidate paths $K$
\Ensure Answer $y \in \mathcal{Y}$

\State \textbf{// Step 1: Meta Operation Decomposition (MOD)}
\State Extract table metadata $\mathcal{T}_{meta}$ from $T$
\State $\mathcal{M}_q \gets$ MOD($q$, $\mathcal{T}_{meta}$) \Comment{Set of high-level meta operations}

\State \textbf{// Step 2: Candidate Macro Path Construction}
\State Initialize candidate path set $\Pi \gets \{\}$ 
\For{$k = 1$ to $K$}
    \State $\pi_k \gets$ generate\_candidate\_path($\mathcal{M}_q$) \Comment{Ordered sequence of meta operations}
    \State $\Pi \gets \Pi \cup \{\pi_k\}$
\EndFor

\State \textbf{// Step 3: Iterative Planning and Execution}
\While{not all paths executed or converged}
    \For{each path $\pi \in \Pi$}
        \State Compute expectation score $\mathbb{E}(\pi)$: 
        \[
        \mathbb{E}(\pi) = \hat{R}(\pi) + c \cdot P(\pi) \sqrt{\frac{\log \sum_{\pi'} N(\pi')}{1 + N(\pi)}}
        \]
    \EndFor
    \State Select top-ranked path(s) $\pi^*$ according to $\mathbb{E}(\pi)$
    \State $context \gets$ initial execution state in $\mathbb{E}$
    
    \For{each meta operation $m$ in $\pi^*$}
        \State $o, r \gets$ execute($m$, $context$, $\mathcal{E}$) \Comment{Execute and observe outcome and reward}
        \State Update Execution Experience Memory: $\mathcal{D} \gets \mathcal{D} \cup \{(m, context, o, r)\}$
        \State Update path reward:
        \[
        \hat{R}(\pi^*) \gets \frac{N(\pi^*) \hat{R}(\pi^*) + r}{N(\pi^*) + 1}, \quad N(\pi^*) \gets N(\pi^*) + 1
        \]
        \If{execution failed or outcome invalid}
            \State Replan: $\pi^* \gets$ Macro Path Planner($\mathcal{M}_q$, $\mathcal{D}$)
            \State break \Comment{Restart path execution with revised plan}
        \EndIf
        \State Update $context$ according to $o$
    \EndFor
\EndWhile

\State \textbf{// Step 4: Return final answer}
\State $y \gets$ extract\_answer($\pi^*$, $\mathcal{D}$)
\State \Return $y$
\end{algorithmic}
\end{algorithm}

\section{DTR-Bench Dataset Curation}
We introduce \textbf{DTR-Bench} (Deep Tabular Research Benchmark), a specialized benchmark for evaluating deep analytical reasoning capabilities over tabular data. Unlike existing table QA benchmarks that focus on simple fact retrieval or single-step reasoning, DTR-Bench emphasizes complex statistical analysis tasks that reflect real-world data science workflows. The benchmark comprises \textbf{500 scenario-driven question-answer pairs} from diverse domains, each requiring sophisticated analytical reasoning such as correlation analysis, inequality measurement, anomaly detection, and statistical hypothesis testing. 
This benchmark is derived from a curated selection of Excel spreadsheets sourced from RealHitBench, encompassing various domains including Economy, Business, and Education.

\subsection{Scenario-Driven Question Generation}
\label{sec:question_generation}

A key innovation of DTR-Bench is the \textbf{scenario-driven question generation} approach. Each question is grounded in a realistic user persona with domain-specific analytical needs, ensuring that questions reflect authentic data science tasks rather than artificial academic exercises.

\paragraph{User Persona Design}
\label{sec:personas}

We define 8 distinct user personas, each representing a real-world role that regularly performs deep tabular analysis:

\begin{table}[h]
\centering
\caption{User Persona Distribution in DTR-Bench}
\label{tab:persona_distribution}
\resizebox{\textwidth}{!}{%
\begin{tabular}{lrl}
\toprule
\textbf{User Persona} & \textbf{\ Questions} & \textbf{Primary Analysis Focus} \\
\midrule
Social Researcher & 105 & Inequality analysis (Gini coefficient, Lorenz curves), demographic patterns \\
Student/Researcher & 88 & Statistical hypothesis testing (ANOVA), effect size calculation (Cohen's $d$, $\eta^2$) \\
Government Staff & 62 & Longitudinal policy impact assessment, equity-efficiency trade-offs \\
Data Analyst & 61 & Dimensionality reduction, cohort analysis, z-score-based anomaly detection \\
Business Owner & 55 & Profitability decomposition, BCG matrix classification, break-even analysis \\
Procurement Manager & 53 & Supplier concentration (HHI), risk-performance matrices \\
Sales Manager & 40 & Multi-dimensional performance analysis, trend detection \\
Investor & 36 & Risk-return analysis, efficient frontier identification, valuation ratios \\
\bottomrule
\end{tabular}%
}
\end{table}

\paragraph{Question Template Design}
Each persona has 4 specialized question templates requiring analytical reasoning. Templates are mainly designed to: a. \textbf{Reference specific columns} using quoted identifiers (e.g., \texttt{"Revenue"})
b. \textbf{Specify analytical methods} (correlation, ANOVA, Gini coefficient, HHI, etc.) c.  \textbf{Request structured outputs} (rankings, statistical measures, interpretations). Table~\ref{tab:persona-templates} illustrates representative question templates for each persona type.

\begin{table}[htbp]
\centering
\caption{Representative Question Templates by Persona}
\label{tab:persona-templates}
\small
\resizebox{\textwidth}{!}{%
\begin{tabular}{@{}p{2.8cm}p{2.8cm}p{10cm}@{}}
\toprule
\textbf{Persona} & \textbf{Analysis Type} & \textbf{Template Example} \\
\midrule
Investor & Risk-Return Analysis & 
Conduct a risk-return analysis across all \texttt{\{Category\}} by examining the relationship between \texttt{\{NumericCol1\}} (returns) and \texttt{\{NumericCol2\}} (risk metrics). Identify the efficient frontier segments. \\
\addlinespace[3pt]
Social Researcher & Inequality Analysis & 
Measure the distribution of \texttt{\{NumericCol\}} across \texttt{\{Category\}} using Gini coefficient and decile ratios. Investigate structural factors driving inequality. \\
\addlinespace[3pt]
Data Analyst & Anomaly Detection & 
Use statistical methods (z-scores) to identify unusual patterns in \texttt{\{NumericCol\}} across \texttt{\{Category\}}. Investigate anomalies and assess impact. \\
\addlinespace[3pt]
Student/Researcher & Statistical Testing & 
Conduct rigorous statistical analysis of \texttt{\{NumericCol\}} across \texttt{\{Category\}}: test for normality, perform ANOVA/Kruskal-Wallis tests, and calculate effect sizes. \\
\bottomrule
\end{tabular}
} 
\end{table}


\subsection{Quality Assurance}
\label{sec:quality}
To ensure benchmark quality, we implement multiple validation steps:
\begin{itemize}
    \setlength{\itemsep}{0pt}
    \setlength{\parskip}{0pt}
    \setlength{\parsep}{0pt}
    \item \textbf{Column Validity Filtering}: Columns with invalid names (e.g., ``Unnamed:'', purely numeric headers, CJK characters) are excluded
    \item \textbf{Template-Table Compatibility}: Templates requiring specific column types (numeric, categorical, temporal) are only instantiated when suitable columns exist
    \item \textbf{Time-Series Validation}: Templates involving trend analysis are skipped for tables without detectable temporal columns
    \item \textbf{Duplicate Prevention}: Questions are deduplicated using normalized string comparison
    \item \textbf{Answer Verification}: All reference answers are computed programmatically from actual table data, ensuring reproducibility
\end{itemize}

\paragraph{Answer Key Points (KeyPoints) for Evaluation}
\label{sec:keypoints}
In addition to validating that reference answers are computed from the underlying table data, we attach an \texttt{AnswerKeyPoints} field to each instance as explicit, machine-checkable grading criteria. This enables fine-grained evaluation (e.g., key-point coverage) beyond string matching.

\begin{figure}[h]
\begin{tcolorbox}[
    colback=gray!5,
    colframe=gray!60,
    fonttitle=\bfseries,
    boxrule=0.5pt
]
\textbf{Answer:} \\
    Gini coefficient of total Revenue across Region: 0.342... \\
\textbf{AnswerKeyPoints:} \\
    "Aggregate the numeric metric by category.", \\
    "Compute an inequality metric such as the Gini coefficient.", \\
    "Report concentration (e.g., top vs bottom shares) and top categories."
\end{tcolorbox}
\end{figure}

\section{Implementation Details}\label{implement}
\subsection{Benchmark Datasets}
\label{sec:datasets}

\paragraph{DTR-Bench: Long-Horizon Analytical Queries.}
To profoundly evaluate the performance for complex and long-horizon tabular reasoning datasets, we construct an additional DTR-Bench based on tables in RealHitBench. Using table meta information and expert-designed templates combined with \texttt{DeepSeek-3.2}, we generate 500 long-form analytical queries that require multi-step reasoning and execution. These queries span five categories: \textit{Analysis}: multi-stage aggregation and interpretation. \textit{Visualization}: chart generation and visual comparison. \textit{Calculation}: chained numerical computation. \textit{Comparison}: cross-group or temporal comparison. \textit{Conditional Calculation}: conditional aggregation and filtering.

Each query typically requires planning over multiple operations and intermediate execution states, making them unsuitable for shallow or single-pass reasoning approaches.

For the evaluation, we report two following metrics. \textit{Win Rate}: proportion of instances where a model outperforms baselines (ties counted as $0$). \textit{Score Rate}: proportion of instances where a model is not worse than baselines (ties counted as $0.5$).

\paragraph{RealHitBench.}
We adopt RealHitBench~\cite{wu2025realhitbench} as our primary benchmark, a large-scale dataset designed for evaluating reasoning over real-world, unstructured tables.
RealHitBench contains tables with heterogeneous layouts, including merged cells, bidirectional and hierarchical headers, missing values, and implicit semantic regions.
The benchmark categorizes tasks into multiple reasoning types and provides fine-grained evaluation protocols.

We evaluate DTR on the following RealHitBench task categories: \textit{Fact Checking}: verifying factual statements grounded in table entries. \textit{Numerical Reasoning}: performing arithmetic and aggregation over table values. \textit{Structure Comprehension}: understanding table organization, header hierarchy, and alignment. \textit{Data Analysis}: multi-step analytical reasoning requiring aggregation, comparison, and synthesis. \textit{Visualization}: generating charts or structured visual outputs based on tabular data.

For Fact Checking, Numerical Reasoning, and Structure Comprehension, we report \textit{Exact Match (EM)} and \textit{F1} scores following the benchmark protocol.
Data Analysis tasks are evaluated using \textit{LLM-based evaluation} and \textit{ROUGE} metrics to assess semantic correctness and completeness.
Visualization tasks are evaluated using \textit{Execution Correctness Rate (ECR)} and \textit{Pass@1}, measuring whether the generated visualization code executes successfully and produces the expected output.

\subsection{Baselines}
\label{sec:baselines}

We compare DTR against a diverse set of baselines spanning table-specialized models, general-purpose large language models, and agent-based reasoning frameworks~\cite{zheng2024multimodal, lu2025youtu}. 
\textbf{Table-specific Models.} TableGPT~\cite{gong2020tablegpt}: a representative table-focused language model designed for structured table question answering. \textbf{General-purpose LLMs.} DeepSeek-3.2: a strong open-source large language model with competitive reasoning capabilities; \textbf{Agentic Frameworks.} ST-RAPTOR~\cite{tang2025st}: a retrieval-augmented agent framework for structured reasoning; Tree Thinker~\cite{wu2025realhitbench}: a tree-based reasoning agent that explores multiple reasoning branches; \textbf{CodeLoop}: we craft a straightforward execution-centric agentic framework that iteratively generates and debugs code. All baselines are evaluated under comparable settings, with access to the same table inputs and query information.
Unless otherwise specified, we use publicly released model checkpoints and default decoding configurations. The base LLMs we adopted is Qwen3 1.7B\&4B~\cite{yang2025qwen3} and DeepSeek V3~\cite{liu2024deepseek}.

\section{Related Work}
\subsection{Deep Research and Agentic Reasoning}
Recent advances in large language models have aroused growing interest in agentic systems that perform multi-step reasoning through interaction with external tools, environments, or intermediate states~\cite{zhao2024expel,huang2024understanding,gong2020tablegpt}. A line of deep research work often focuses on long-horizon problem solving, where models iteratively plan and revise their strategies based on intermediate feedback~\cite{wolflein2025llm}. Typical examples include agents augmented with external tools, systems that separate planning from execution, and methods that rely on self-reflection or self-correction across multiple steps~\cite{renze2024self,dagan2023dynamic}. Notably, many existing agent-based approaches emphasis on language-level planning and reasoning~\cite{ge2025samule,song2023llm}. While such approaches have demonstrated strong performance on structured benchmarks, they typically assume reliable intermediate steps and make limited use of execution feedback. In contrast, our work treats deep search as a continual process, where accumulated action experience guides path selection and enables more robust behavior.

\subsection{Tabular Reasoning and Table Question Answering}

Tabular reasoning has been extensively studied in the context of table question answering, semantic parsing, and data analysis~\cite{zhao2023openrt}. Early approaches typically focused on mapping natural language queries to logical forms or executable programs over well-structured tables with clean schemas and regular layouts~\cite{akhtar2023exploring,Perzina2014MicrosoftEA}. Recent methods leverage large language models to perform tabular reasoning in a more flexible manner through straightforwardly interpreting table contents. Such pipelines reason over tables using code generation, textual serialization, or prompting strategies~\cite{wang2024chain,Zhu2024AutoTQATA,Shi2024EHRAgentCE}. Existing work on tabular reasoning spans a diverse set of tasks, including fact checking, numerical reasoning, data analysis, and table-based visualization~\cite{Wu2024TableBenchAC}, where models are required to interpret structured or semi-structured information and perform reasoning over rows, columns, and cell values~\cite{Sun2024ASO}. Despite substantial progress, chain-of-thought prompting and single-pass code generation can be sensitive to early reasoning decisions in complex tabular analysis. While hybrid symbolic approaches improve robustness, they lack mechanisms for iterative decision making with accumulated experience~\cite{kulkarni2025llmsymbolicintegrationrobusttemporal}. 

Our work bridges this gap by introducing a unified framework for Deep Tabular Research that integrates operator-level execution with continual experience-driven decision making. By explicitly modeling path selection, execution, and memory as interacting components, the proposed approach extends prior reasoning systems to handle complex, unstructured tables requiring iterative analysis and revision.


\section{Case Studies: Real Trajectories}
\subsection{Case Study 1: Pie Chart - Employment Distribution}

\subsubsection{Task}

\textbf{Question}: Using the data, draw a pie chart to illustrate the distribution of employment between agriculture and non-agriculture sectors in 1984.

\textbf{Question Type}: Visualization

\textbf{Table Context}: Employment statistics table with Year, Agriculture employment, and Nonagricultural industries columns.

\subsubsection{Generated Code}

import pandas as pd \\
import matplotlib.pyplot as plt \\
\\
 Extract data for 1984 (row index 5) \\
agriculture = 2668 \\
non\_agriculture = 56423 \\
\\
 Prepare data for pie chart \\
sectors = ['Agriculture', 'Non-Agriculture'] \\
employment = [agriculture, non\_agriculture] \\
\\
 Verify extraction \\
print("Sectors:", sectors) \\
print("Employment:", employment) \\
assert len(employment) > 0, "Employment data cannot be empty!" \\
assert all(isinstance(e, (int, float)) for e in employment), \ \\
      "Employment values must be numbers!" \\
\\
 Create pie chart \\
plt.figure(figsize=(8, 8)) \\
plt.pie(employment, labels=sectors, autopct='\%1.1f\%\%', \\
        startangle=90, colors=['ff9999','66b3ff']) \\
plt.title('Employment Distribution in 1984 ' + \\
          '(Agriculture vs Non-Agriculture)') \\
plt.tight\_layout() \\
plt.show() \\
\\

\subsubsection{Generated Visualization}

\begin{figure}[H]
\centering
\includegraphics[width=0.5\textwidth]{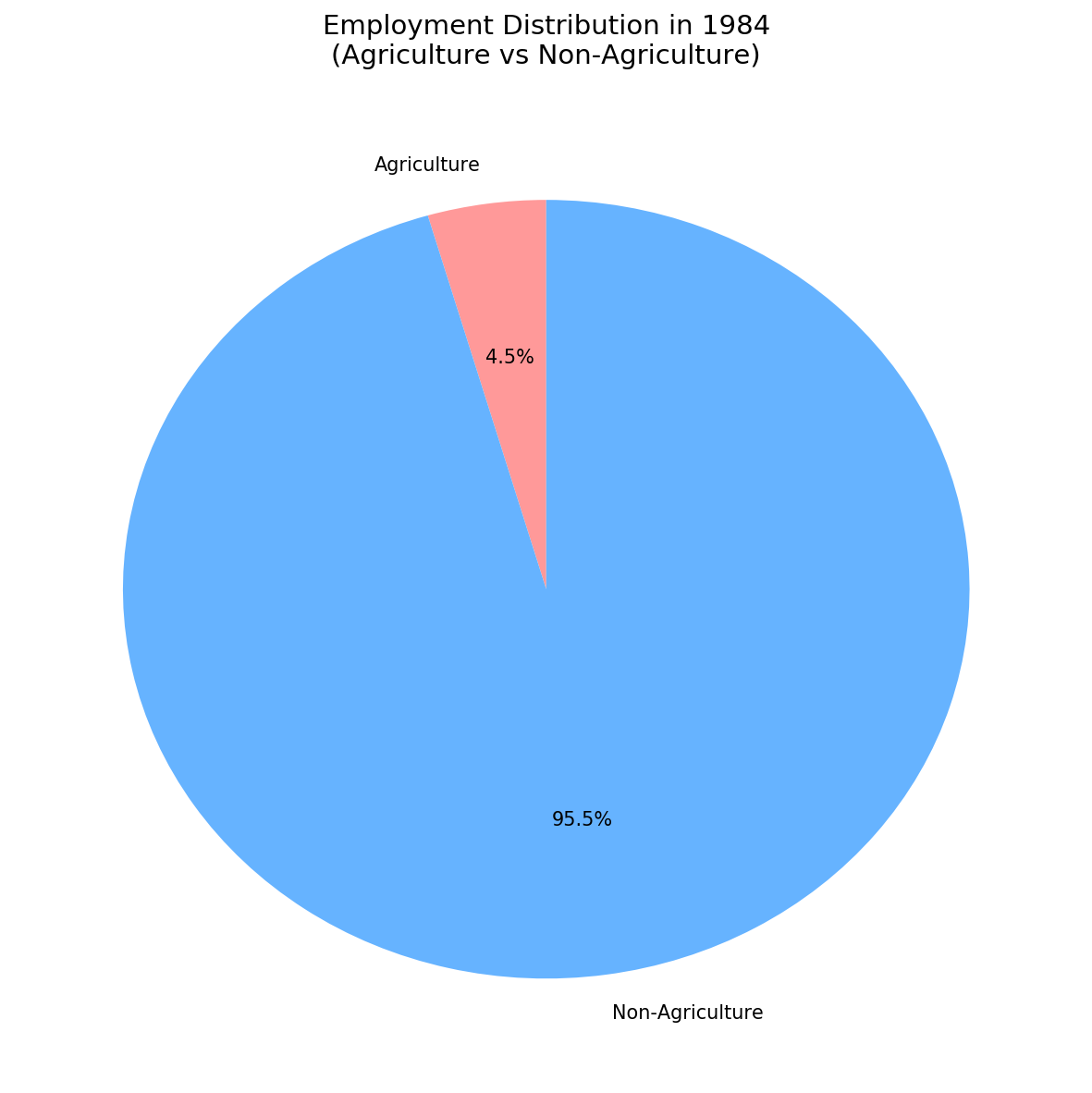}
\caption{Employment Distribution in 1984: Agriculture (4.5\%) vs Non-Agriculture (95.5\%)}
\end{figure}

\subsubsection{Quality Analysis}

\textbf{Strengths}:
\begin{itemize}
    \item \checkmark Correctly extracted data from 1984 row
    \item \checkmark Clean, executable code with data validation (assertions)
    \item \checkmark Appropriate visualization aesthetics (colors, sizing, labels)
    \item \checkmark Added percentage labels using \texttt{autopct} for clarity
    \item \checkmark Clear title explaining what the chart shows
\end{itemize}

\textbf{Key Insight}: The chart clearly shows non-agricultural employment dominated 1984 with 95.5\%, while agriculture accounted for only 4.5\%, illustrating the economic transformation towards industrial and service sectors.

\newpage

\subsection{Case Study 2: Line Chart - Employment Trends}

\subsubsection{Task}

\textbf{Question}: Please help me create a line chart to compare the trends of total employed population and total unemployed population from 1983 to 1990.

\textbf{Question Type}: Visualization

\textbf{Table Context}: Employment statistics with yearly data including total employed and unemployed counts.

\subsubsection{Generated Code}

import pandas as pd \\
import matplotlib.pyplot as plt \\
\\
 Extract data from table \\
years = [1983, 1984, 1985, 1986, 1987, 1988, 1989, 1990] \\
employed = [56787, 59091, 59891, 60892, 62107, 63273, 64315, 65104] \\
unemployed = [6260, 4744, 4521, 4530, 4101, 3655, 3525, 3906] \\
\\
 Verify extraction \\
print("Years:", years) \\
print("Employed:", employed) \\
print("Unemployed:", unemployed) \\
assert len(years) == len(employed) == len(unemployed), \ \\
      "Data length mismatch!" \\
assert all(isinstance(y, (int, float)) for y in employed + unemployed), \ \\
      "Values must be numbers!" \\
\\
 Create visualization \\
plt.figure(figsize=(10, 6)) \\
plt.plot(years, employed, marker='o', label='Employed') \\
plt.plot(years, unemployed, marker='o', label='Unemployed') \\
\\
plt.xlabel('Year') \\
plt.ylabel('Population (thousands)') \\
plt.title('Employment Trends (1983-1990)') \\
plt.legend() \\
plt.grid(True) \\
plt.tight\_layout() \\
plt.show() \\
\\

\subsubsection{Generated Visualization}

\begin{figure}[H]
\centering
\includegraphics[width=0.9\textwidth]{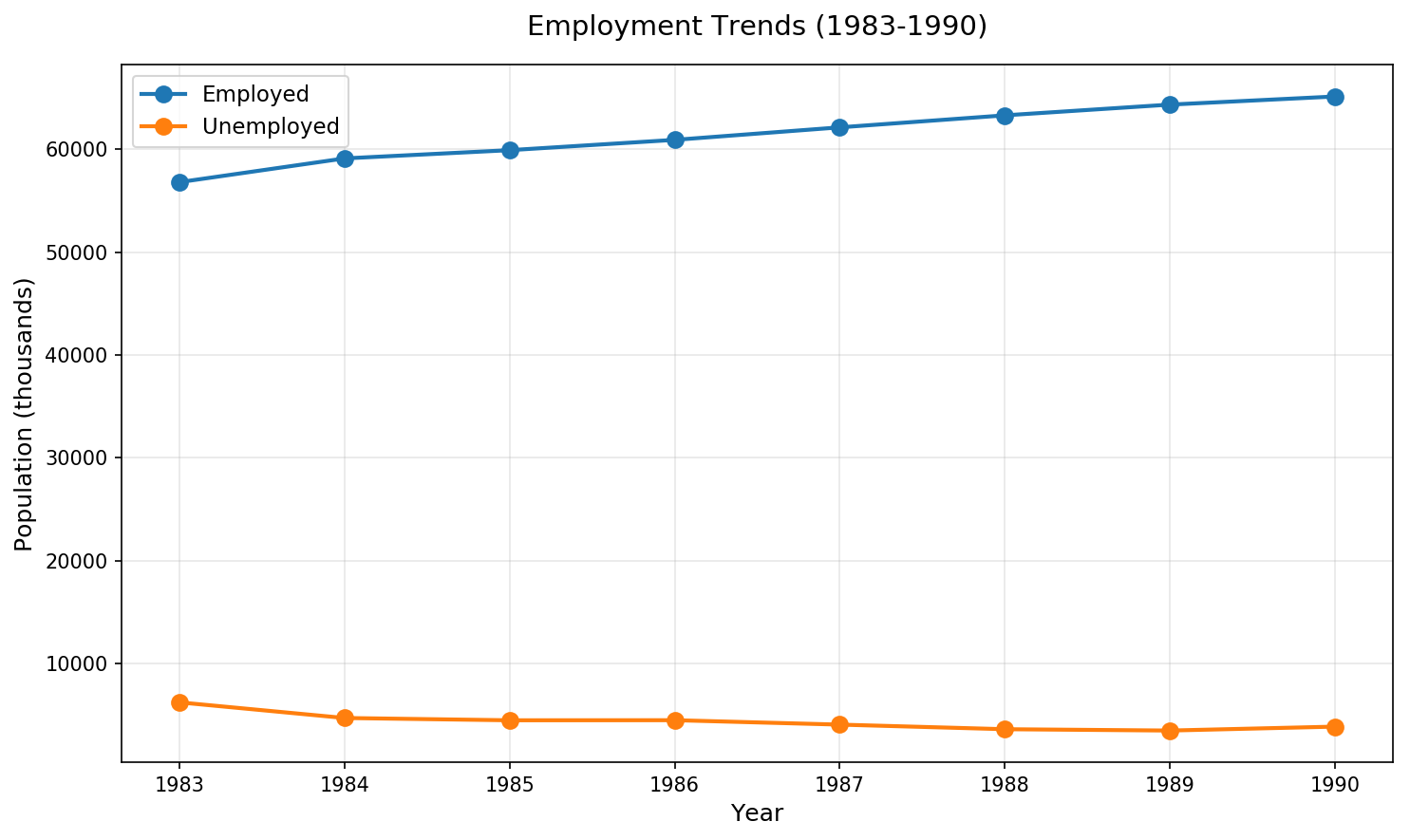}
\caption{Employment Trends 1983-1990: Employed population grew from 56.8M to 65.1M while unemployed declined from 6.3M to 3.9M}
\end{figure}

\subsubsection{Analysis}

\textbf{Data Extraction Quality}:
\begin{itemize}
    \item Accurately extracted 8-year time series across multiple columns
    \item Comprehensive data validation with assertions
    \item Proper data type checking
\end{itemize}

\textbf{Visualization Design}:
\begin{itemize}
    \item Different series with markers for clarity
    \item Grid enabled for easier value reading
    \item Proper legend placement
    \item Clear axis labels with units (thousands)
\end{itemize}

\textbf{Economic Insights}:
\begin{itemize}
    \item Chart reveals steady employment growth (56.8M $\to$ 65.1M, +14.6\%)
    \item Declining unemployment (6.3M $\to$ 3.9M, -37.6\%)
    \item Indicates strong economic recovery period during 1980s
    \item Unemployment rate dropped from ~10\% to ~5.6\% over this period
\end{itemize}

\newpage

\subsection{Case Study 3: Scatter Plot - Correlation Analysis}

\subsubsection{Task}

\textbf{Question}: Create a scatter plot to visualize the relationship between the unemployment rate and the total civilian labor force for all years in the dataset.

\textbf{Question Type}: Visualization + Data Analysis

\textbf{Table Context}: Long-term employment statistics (1983-1998) with civilian labor force size and unemployment rates.

\subsubsection{Generated Code}

import pandas as pd \\
import matplotlib.pyplot as plt \\
\\
 Extract data from table \\
x\_values = [1983, 1984, 1985, 1986, 1987, 1988, 1989, 1990, \\
            1991, 1992, 1993, 1994, 1995, 1996, 1997, 1998] \\
y\_values = [9.9, 7.4, 7.0, 6.9, 6.2, 5.5, 5.2, 5.7, \\
            7.2, 7.9, 7.2, 6.2, 5.6, 5.4, 4.9, 4.4] \\
civilian\_labor\_force = [63047, 63835, 64411, 65422, 66207, 66927, \\
                        67840, 69011, 69168, 69964, 70404, 70817, \\
                        71360, 72087, 73261, 73959] \\
\\
 Verify extraction \\
print("Years:", x\_values) \\
print("Unemployment rates:", y\_values) \\
print("Civilian labor force:", civilian\_labor\_force) \\
assert len(y\_values) > 0, "Y values cannot be empty!" \\
assert all(isinstance(y, (int, float)) for y in y\_values), \ \\
      "Y values must be numbers!" \\
\\
 Create visualization \\
plt.figure(figsize=(12, 7)) \\
plt.scatter(civilian\_labor\_force, y\_values, c='blue', alpha=0.7) \\
\\
plt.xlabel('Civilian Labor Force (thousands)') \\
plt.ylabel('Unemployment Rate (\%)') \\
plt.title('Relationship Between Unemployment Rate and Civilian ' + \\
          'Labor Force (1983-1998)') \\
plt.grid(True) \\
plt.tight\_layout() \\
plt.show() \\
\\

\subsubsection{Generated Visualization}

\begin{figure}[H]
\centering
\includegraphics[width=0.95\textwidth]{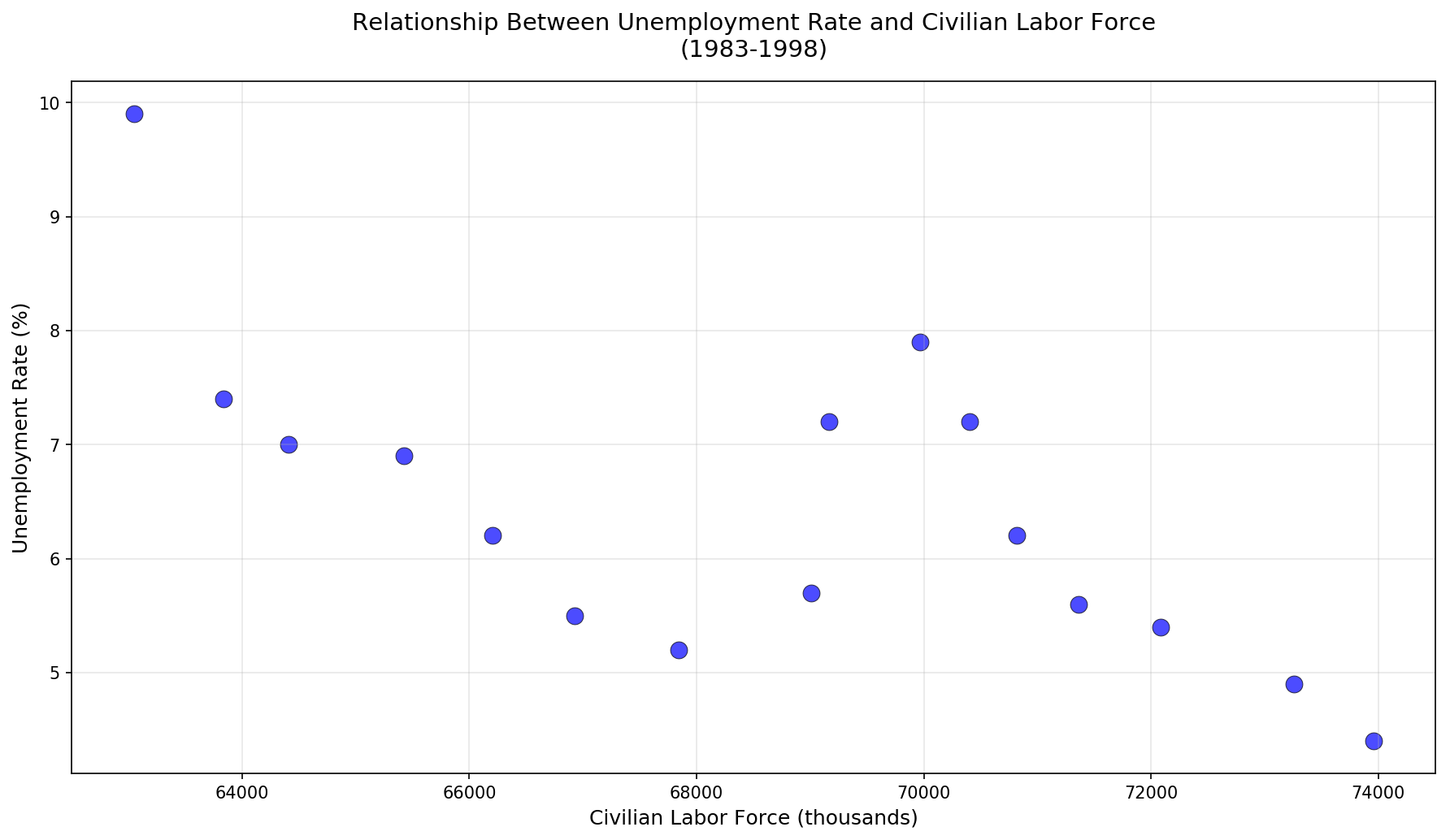}
\caption{Scatter plot showing the relationship between unemployment rate and civilian labor force size over 16 years (1983-1998)}
\end{figure}

\subsubsection{Advanced Analysis}

\textbf{Multi-variable Extraction}:
\begin{itemize}
    \item Successfully extracted three related data series
    \item 16 years of continuous data (1983-1998)
    \item Proper alignment between years, rates, and labor force size
\end{itemize}

\textbf{Correlation Insights}:
\begin{itemize}
    \item \textbf{Inverse Relationship}: As labor force grows, unemployment tends to decrease
    \item High unemployment (9.9\%) at lower labor force size (63M) in 1983
    \item Unemployment stabilizes around 4-6\% as labor force expands to 74M
    \item Demonstrates economic expansion pattern: growing labor force absorption
\end{itemize}

\textbf{Economic Interpretation}:
\begin{itemize}
    \item \textbf{1983-1990}: Steep decline from 9.9\% to 5.2\% unemployment despite labor force growth of 9.4\% - indicates strong job creation
    \item \textbf{1990-1992}: Unemployment spike to 7.9\% (recession period)
    \item \textbf{1993-1998}: Steady improvement to 4.4\% with continued labor force expansion - economic boom period
\end{itemize}

\textbf{Code Features}:
\begin{itemize}
    \item Alpha transparency (0.7) for better point visibility
    \item Appropriate figure sizing (12x7) for wide scatter plots
    \item Grid enabled for easier coordinate reading
    \item Comprehensive data validation before plotting
\end{itemize}

\newpage

\section{Data Analysis Case Study}

\subsection{Case Study 4: Comprehensive Statistical Analysis}

\subsubsection{Task}

\textbf{Question Type}: Data Analysis (Rudimentary Analysis)

\textbf{Question}: What is the average unemployment rate across all years, and what is its standard deviation?

\textbf{Table Context}: 71 years of unemployment data (1953-2023)

\textbf{Expected Answer}: Mean = 5.80\%, Std = 1.62\%

\subsubsection{Generated Code and Analysis}

import pandas as pd \\
import numpy as np \\
\\
 Extract unemployment rate data \\
unemployment\_rates = df['Percent of labor force'].values \\
\\
 Calculate statistics \\
mean\_rate = np.mean(unemployment\_rates) \\
std\_rate = np.std(unemployment\_rates, ddof=1) \\
\\
 Create result DataFrame \\
df = pd.DataFrame({ \\
    'Mean': [round(mean\_rate, 2)], \\
    'Std Dev': [round(std\_rate, 2)] \\
}) \\
\\
print(f"Average unemployment rate: {mean\_rate:.2f}
print(f"Standard deviation: {std\_rate:.2f}
\\

\subsubsection{Model Response (Detailed Analysis)}

\begin{quote}
\textbf{Data Overview:}\\
Analyzed unemployment rates across 72 years (1953-2023). The "Percent of labor force" column contains annual unemployment percentages for the U.S. civilian labor force.

\textbf{Calculation Process:}
\begin{enumerate}
    \item Extracted all 72 values from "Percent of labor force" column
    \item Calculated mean: Sum(411.5) / Count(71) = 5.80\%
    \item Calculated std: $\sqrt{\frac{\Sigma(x\_i - \bar{x})^2}{n-1}} = 1.62\%$
\end{enumerate}

\textbf{Statistical Details:}
\begin{itemize}
    \item Mean: 5.80\%
    \item Standard Deviation: 1.62\%
    \item Minimum: 2.9\% (1953)
    \item Maximum: 9.7\% (1982)
    \item Range: 6.8 percentage points
    \item Sample size: 71 valid data points
\end{itemize}

\textbf{Insights:}\\
The average unemployment of 5.80\% represents typical U.S. unemployment over 70 years. Standard deviation of 1.62\% indicates moderate variability, with most years in 4.2-7.4\% range ($\pm1\sigma$). Higher values occurred during recessions (1982, 2010, 2020).

\textbf{Context:}\\
Peak unemployment during 1982 recession (9.7\%) and 2008 financial crisis (9.6\% in 2010). Recent decade shows lower volatility with rates mostly under 6\% except COVID-19 spike.
\end{quote}

\subsubsection{Quality Indicators}

\begin{itemize}
    \item \checkmark Correct numerical results (5.80\%, 1.62\%)
    \item \checkmark Detailed calculation methodology with formulas
    \item \checkmark Rich contextual analysis with historical events
    \item \checkmark Statistical interpretation ($\pm1\sigma$ range)
    \item \checkmark Identification of outliers and anomalies
    \item \checkmark Time-series context (70+ years of data)
\end{itemize}

\end{document}